\pdfoutput=1

\documentclass[11pt]{article}

\usepackage[]{acl}

\usepackage{times}
\usepackage{latexsym}

\usepackage[T1]{fontenc}

\usepackage[utf8]{inputenc}
\usepackage{xcolor}

\usepackage{microtype}
\usepackage{inconsolata}
\usepackage[linesnumbered,ruled,vlined]{algorithm2e}
\usepackage{graphicx}
\usepackage{bm}
\usepackage{enumitem}
\usepackage{multirow}
\usepackage{amsfonts}
\usepackage{amssymb}
\usepackage{amsmath}
\usepackage{verbatim}
\usepackage{booktabs}
\usepackage{arydshln}
\usepackage{wasysym}

\SetKwInput{KwInput}{Input}                
\SetKwInput{KwOutput}{Output}

\usepackage{cleveref}
\newcommand{\metricname}[1]{\textsc{Familiarity}}

\title{\metricname{}: Better Evaluation of Zero-Shot Named Entity Recognition by Quantifying Label Shifts in Synthetic Training Data}

\author{
    Jonas Golde\textsuperscript{1}, Patrick Haller\textsuperscript{1}, Max Ploner\textsuperscript{1}, \\\vspace{1mm} \textbf{Fabio Barth\textsuperscript{2},} \textbf{Nicolaas Jedema\textsuperscript{3},} \textbf{Alan Akbik\textsuperscript{1}} \\
    \textsuperscript{1}Humboldt Universität zu Berlin, \textsuperscript{2}DFKI, \textsuperscript{3}Amazon \\
}

\begin{document}
\maketitle
\begin{abstract}
Zero-shot named entity recognition (NER) is the task of detecting named entities of specific types (such as \textsc{Person} or \textsc{Medicine}) without any training examples. Current research increasingly relies on large synthetic datasets, automatically generated to cover tens of thousands of distinct entity types, to train zero-shot NER models. 
However, in this paper, we find that these synthetic datasets often contain entity types that are semantically highly similar to (or even the same as) those in standard evaluation benchmarks. Because of this overlap, we argue that reported F1 scores for zero-shot NER overestimate the true capabilities of these approaches. Further, we argue that current evaluation setups provide an incomplete picture of zero-shot abilities since they do not quantify the \textit{label shift} (i.e., the similarity of labels) between training and evaluation datasets. 
To address these issues, we propose \metricname{}, a novel metric that captures both the semantic similarity between entity types in training and evaluation, as well as their frequency in the training data, to provide an estimate of label shift. It allows researchers to contextualize reported zero-shot NER scores when using custom synthetic training datasets. Further, it enables researchers to generate evaluation setups of various transfer difficulties for fine-grained analysis of zero-shot NER.  

\end{abstract}

\section{Introduction} \label{sec:introduction}

\begin{figure}[ht]
\vspace{-2mm}
\includegraphics[width=\columnwidth]{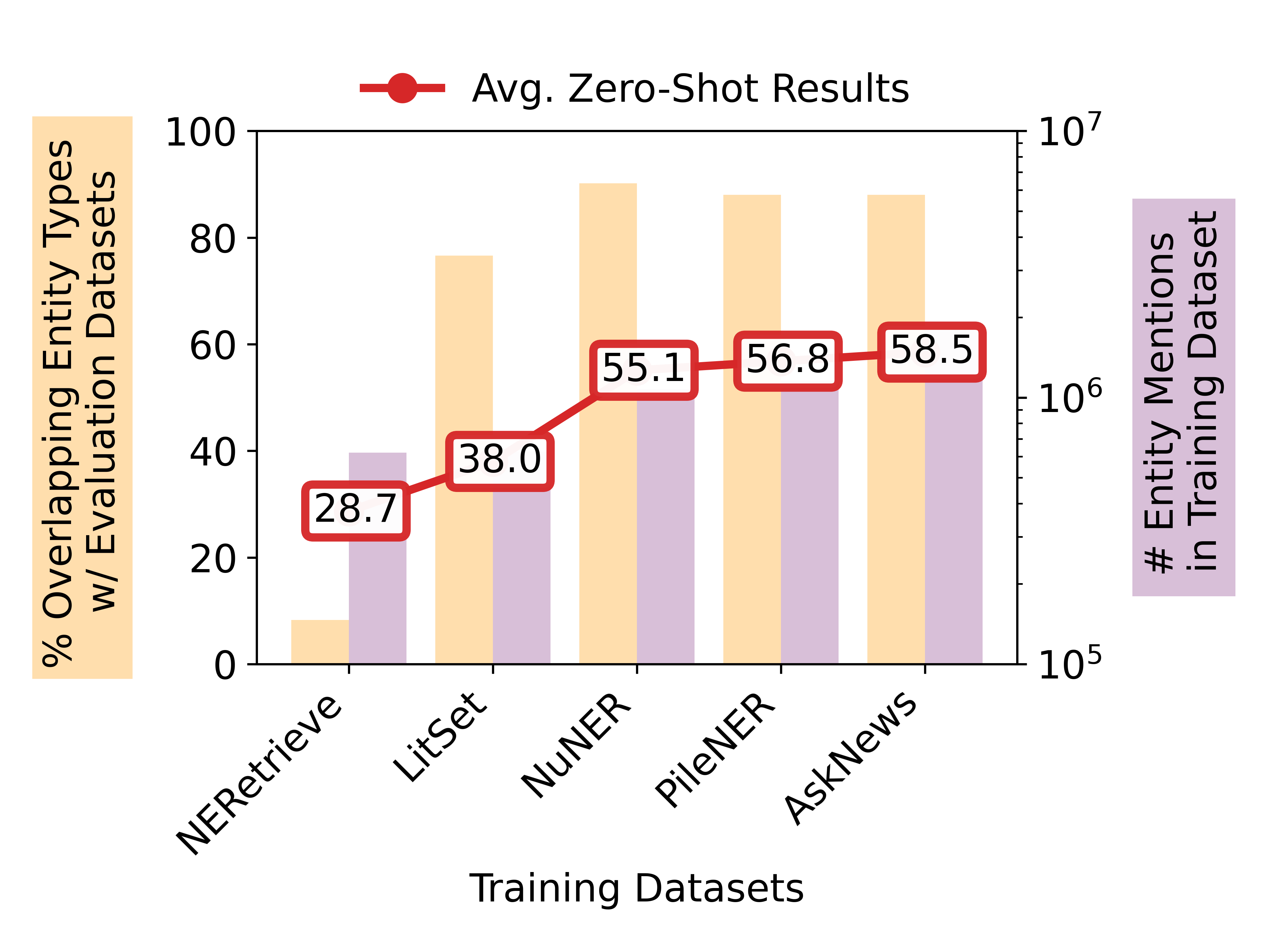}
\vspace{-8mm}
\caption{Impact of training data on zero-shot performance of the current state-of-the-art approach (GLiNER). Each synthetic dataset is characterized by the label overlap (yellow column) and the total number of entity mentions (purple column). While zero-shot performance (red line, macro-averaged F1 across 7 benchmarks) has significantly improved, we note a concerning increase in entity type overlaps between training and testing data.}
\vspace{-3mm}

\label{figure:introduction}
\end{figure}

Zero-shot named entity recognition (NER) is the task of recognizing instances of named entities of specific types (such as \textsc{Person}, \textsc{Organization}, or \textsc{Medicine}) without any training examples. Current state-of-the-art models, such as GLiNER \citep{zaratiana-2023-gliner} and GoLLIE \citep{sainz-2024-gollie}, are initially trained on datasets that contain a large set of different entity types~\citep{aly-etal-2021-leveraging,ma-etal-2022-label}. This allows the models to identify mentions of previously unseen entity types by leveraging their general language understanding capabilities \citep{golde-etal-2024-large}. Finally, these models are evaluated on zero-shot benchmarks that were excluded from the training process \citep{yang-katiyar-2020-simple,das-etal-2022-container,yang-etal-2022-see}.

\begin{figure}[ht]
\includegraphics[width=\columnwidth]{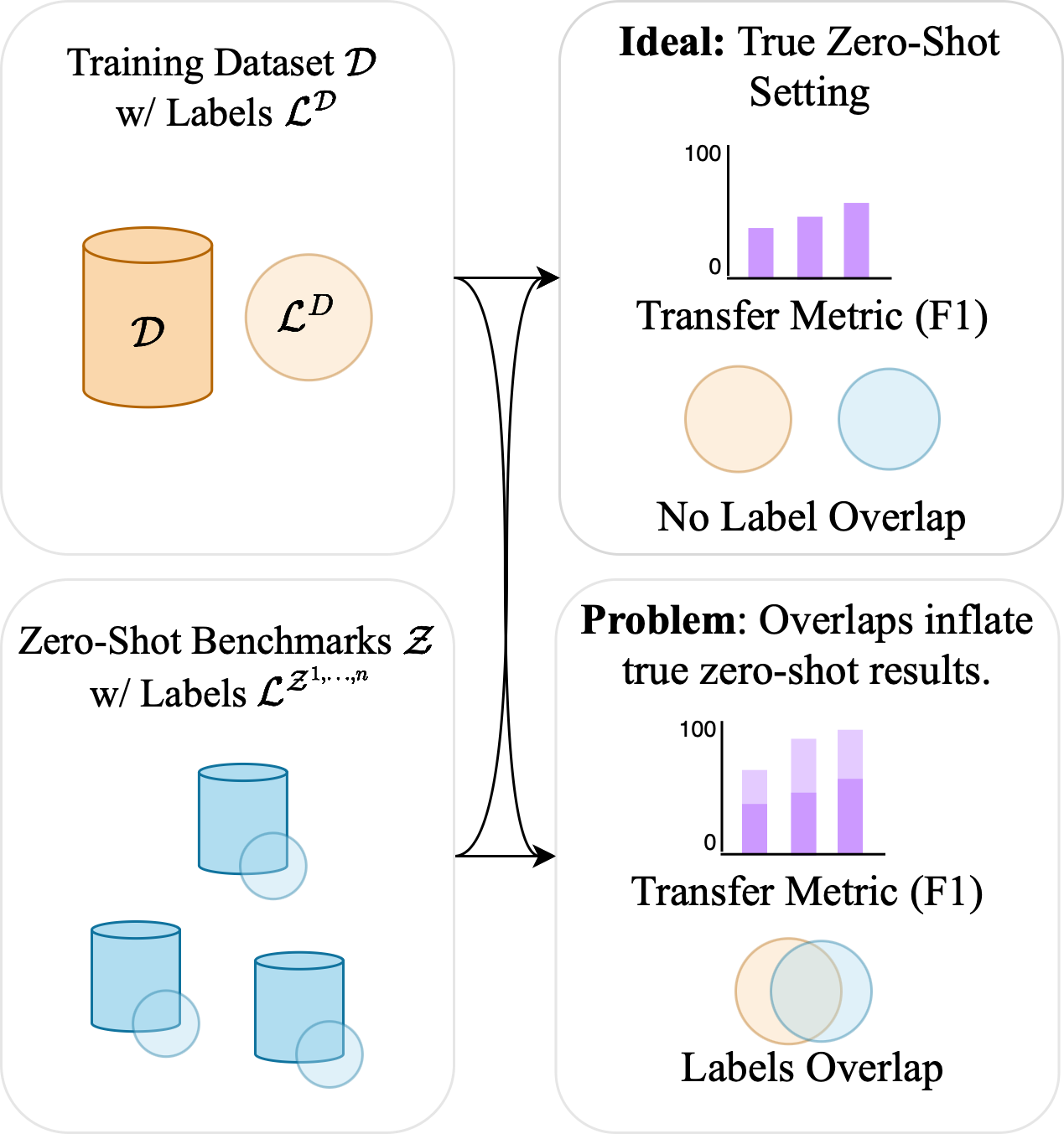}
\caption{With LLMs now capable of generating datasets that cover thousands of entity types, models trained on different datasets are subject to varying label shifts, making comparisons between them challenging. To address this, we introduce \metricname{}, a metric that quantifies and accounts for label shift, enabling more accurate and fair comparisons across models.}
\label{figure:illustration}
\vspace{-4mm}
\end{figure}

\noindent 
\textbf{Advent of large synthetic training datasets.} Recent research has developed methods that can automatically produce training datasets with over tens of thousands of distinct entity types, using available knowledge bases~\citep{wikidata2014} or large language models \citep[LLMs,][]{brown-2020-gpt}. Examples include PileNER \citep{zhou-2023-universalner}, NuNER \citep{bogdanov-2024-nuner}, and AskNews \citep{törnquist2024curatinggroundedsyntheticdata}.  
This represents a paradigm shift for zero-shot NER, which classically relied on hand-labeled training datasets with a much smaller set of entity types, such as Ontonotes~\citep[18 types,][]{hovy-etal-2006-ontonotes}.

As~\Cref{figure:introduction} shows, the advent of large synthetic training datasets has significantly improved reported zero-shot F1 scores. However, as the figure also shows, there is a concerning increase in the overlap between entity types in synthetic datasets and the evaluation benchmarks (cf.~\Cref{figure:introduction}, yellow bars). This means that evaluated models have indeed seen many instances of highly similar (or even the same) entity types during training, raising the question of whether the reported F1 scores overestimate their true zero-shot capabilities. 



\noindent 
\textbf{Broader implications.} Naturally, we could strive to ensure a fair zero-shot comparison by proposing training and evaluation splits that have no overlapping entity types at all. However, ensuring no overlap is in fact not trivial since the same or highly similar entity types might have different labels (such as \textsc{Corporation} and \textsc{Organization}). But more crucially, using fixed training and evaluation splits would potentially limit process driven by advancements in generating synthetic datasets. 

We rather argue that given the advancements of LLMs and their potential to generate high-quality datasets, accepting custom synthetic training datasets is inevitable. We therefore propose to measure the transfer difficulty between the labels of a training and an evaluation dataset, referred to as \textit{label shift}~\citep{lipton2018detecting,wu2021online}.

\noindent 
\textbf{Contributions.} With this paper, we identify a critical issue with current zero-shot NER evaluations caused by the growing availability of large-scale synthetic training datasets. To address this issue, we propose \metricname{}, a novel metric that quantifies the similarity between the sets of entity types in training and evaluation data, allowing us to assess the transfer difficulty of an evaluation setup (cf.~\Cref{figure:illustration}). We summarize our contributions as follows:
\begin{enumerate}
    \item We empirically demonstrate that label overlaps introduce undesirable biases in current zero-shot evaluation setups (\Cref{sec:validation_experiment}).
    \item We propose \metricname{}, a metric that quantifies label shift between training data and evaluation benchmarks, providing insights into transfer difficulty (\Cref{sec:metric}).
    \item We conduct a thorough analysis of \metricname{}, showing that it effectively mitigates the evaluation bias and can be used to generate training splits of varying difficulty levels (\Cref{sec:experiments}).
\end{enumerate}

To enable the research community to efficiently compute \metricname{} and incorporate it into future research, we make all code publicly available as open source\footnote{\url{https://github.com/flairNLP/familiarity}}. Further, we publish three benchmark scenarios on the Hugging Face hub\footnote{\url{https://huggingface.co/flair}} for different levels of transfer difficulty to aid researchers in fine-grained analysis of zero-shot NER.

\section{The Impact of Synthetic Datasets on Current Evaluations} \label{sec:validation_experiment}
As shown in~\Cref{figure:illustration}, we hypothesize that label shift between fine-tuning and evaluation datasets affects transfer performance, particularly in zero-shot NER settings. We define this transfer as the process of fine-tuning a model $\Theta$ on a dataset $\mathcal{D}$ with entity types $\mathcal{L}^{\mathcal{D}}$ and subsequently evaluating it on one or more benchmarks $\mathcal{Z}_{1, \dots, n}$, each with its own set of entity types $\mathcal{L}^{\mathcal{Z}_{1, \dots, n}}$, such that $\mathcal{Z} = \cup_{i=1}^{n} \mathcal{Z}_i$ and $\mathcal{L}^{\mathcal{Z}} = \cup_{i=1}^{n} \mathcal{L}^{\mathcal{Z}_i}$. The datasets themselves do not overlap: $\mathcal{Z} \cap \mathcal{D} = \emptyset$.

However, the entity type sets of the training and evaluation datasets may overlap due to the broad coverage of entity types, particularly in synthetic training datasets: $\mathcal{L}^{\mathcal{Z}} \subseteq \mathcal{L}^{\mathcal{D}}.$

We further note that it is possible that $\mathcal{L}^{\mathcal{Z}} \cap \mathcal{L}^{\mathcal{D}} = \emptyset$. However, given that LLMs can generate fine-tuning datasets with thousands of entity types, we observe that in some cases, more than 80\% of the evaluation entity types are included in the training dataset (e.g., NuNER, PileNER, and AskNews in~\Cref{figure:introduction}). This obviously distorts the genuine zero-shot nature of transfer evaluations, and we hypothesize that the performance for an entity type $\ell$ present in both the evaluation benchmark and the fine-tuning dataset ($\ell \in \mathcal{L}^{\mathcal{Z}} \cap \mathcal{L}^{\mathcal{D}}$) will be higher than for an entity type not present in the fine-tuning data ($\ell \in \mathcal{L}^{\mathcal{Z}} \setminus \mathcal{L}^{\mathcal{D}}$).

\begin{figure*}[t]
\includegraphics[width=\textwidth]{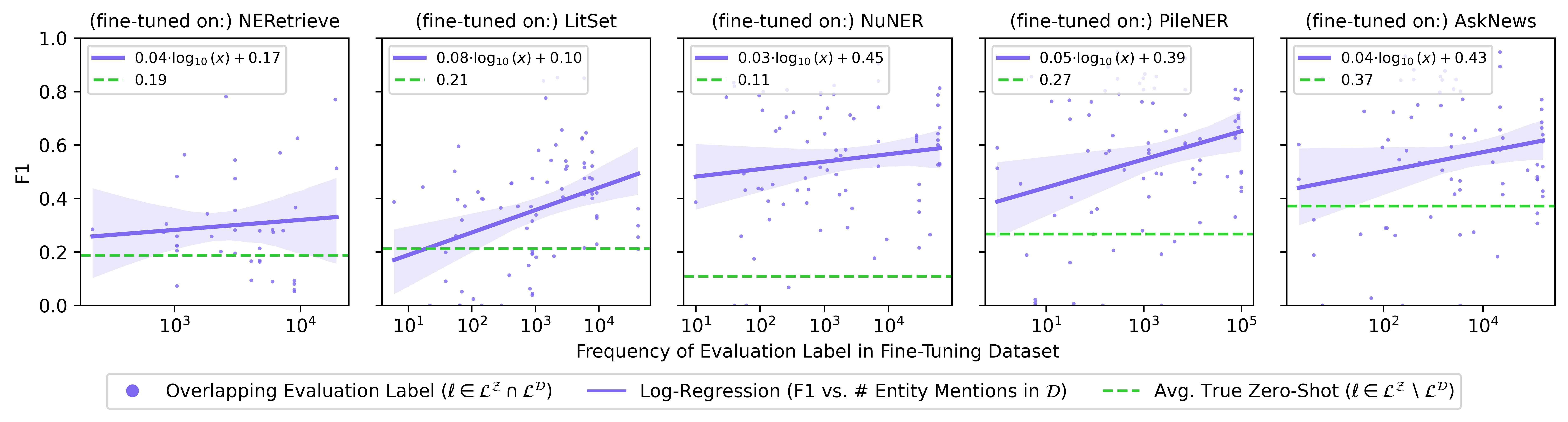}
\caption{Transfer performance is higher on entity types that occur in both evaluation and fine-tuning datasets compared to unseen types. Further, we observe a positive, log-linear correlation between the number of entity mentions for some entity type and its final performance.}
\label{figure:f1_correlation}
\vspace{-4mm}
\end{figure*}

\begin{table}[t]
\centering
\begin{tabular}{lrrr}
\toprule
Dataset & $\#$ Sent. & $\#$ Ent. & $\emptyset$ Ment. \\ 
& & Types & per Sent. \\
\midrule
NERetrieve & 3,437.6k & 0.5k & 2.6 \\
LitSet & 972.6k & 242.9k & 0.8 \\
NuNER & 971.8k & 192.1k & 4.5 \\
PileNER & 45.9k & 12.6k & 20.5 \\
AskNews & 49.4k & 12.6k & 20.2 \\
\bottomrule
\end{tabular}
\caption{Overview of synthetic fine-tuning datasets used in our experiments with their total number of sentences, distinct number of entity types, and average number of entity mentions per sentence.}
\label{table:overview_datasets}
\vspace{-3mm}
\end{table}

\subsection{Experimental Setup}
First, we examine the extent to which label overlaps are a problem, and second, whether synthetic datasets can be scaled to enhance performance through increased examples, considering the potential risk that LLMs may generate duplicate training data, which could lead to performance saturation. To address these questions, we train universal NER models on five large-scale datasets and evaluate them on seven widely used benchmarks. We then analyze the transfer performance for each entity type, classifying them as either overlapping ($\ell \in \mathcal{L}^{\mathcal{Z}} \cap \mathcal{L}^{\mathcal{D}}$) or true zero-shot ($\ell \in \mathcal{L}^{\mathcal{Z}} \setminus \mathcal{L}^{\mathcal{D}}$). For entity types present in both the evaluation and fine-tuning datasets, we perform a log-linear regression to examine whether the number of entity mentions is positively correlated with the performance on those types.

\noindent\textbf{Synthetic fine-tuning datasets.}
We consider five synthetic or automatically derived datasets specifically designed for training zero-shot NER models. NERetrieve \citep{katz-etal-2023-neretrieve} and LitSet \citep{golde-etal-2024-large} are automatically derived from the knowledge bases CaLiGraph \citep{heist2022caligraphontologychallengeowl} and WikiData \citep{wikidata2014}). NuNER \citep{bogdanov-2024-nuner} and PileNER \citep{zhou-2023-universalner} use \texttt{gpt-3.5} \citep{brown-2020-gpt} to annotate large-scale corpora. AskNews \citep{törnquist2024curatinggroundedsyntheticdata} extends NuNER with real-world, diverse news articles obtained from the AskNews API\footnote{\url{https://asknews.app/}}. An overview of these datasets is provided in~\Cref{table:overview_datasets}.

\begin{table}[t]
\centering
\begin{tabular}{lrrr}
\toprule
Dataset & $\#$ Sent. & $\#$ Ent. & $\emptyset$ Ment. \\ 
& & Types & per Sent. \\
\midrule
Movie & 2.4k & 12 & 2.2 \\
Restaurant & 1.5k & 8 & 2.1 \\
AI & 431 & 14 & 4.2 \\
Literature & 416 & 12 & 5.4 \\
Music & 465 & 13 & 7.1 \\
Politics & 650 & 9 & 6.5 \\
Science & 543 & 17 & 5.7 \\
\bottomrule
\end{tabular}
\caption{Overview of the 7 zero-shot benchmarks used in our experiments. Abbreviations are identical to the ones used in~\Cref{table:overview_datasets}.}
\label{table:overview_benchmarks}
\vspace{-3mm}
\end{table}

\noindent\textbf{Zero-shot benchmarks.}
For evaluation, we use the MIT Movie and Restaurant datasets \cite{liu-2013-mit}, as well as the CrossNER dataset \citep{liu-2021-crossner}, as they are frequently used in zero-shot transfer settings \citep{zhou-2023-universalner,zaratiana-2023-gliner,sainz-2024-gollie}. CrossNER includes five domains: Movies, AI, Literature, Politics, and Science. An overview of these datasets is provided in~\Cref{table:overview_benchmarks}.

\noindent\textbf{Training details.}
We use the GLiNER architecture \citep{zaratiana-2023-gliner}, which represents the current state-of-the-art. We reuse all hyperparameters as reported in the original paper. For each of the five datasets, we train a model using three different seeds. To ensure that no model benefits from being trained on significantly more data, we train every model for a fixed number of 60,000 steps with a batch size of 8. The authors of the AskNews model do not train their model from scratch; instead, they continue fine-tuning a model that was initially trained on the NuNER dataset. We follow this approach and further fine-tune our NuNER-trained model for 25 epochs with a batch size of 5, as reported in their paper. We use Hugging Face's Transformers library \citep{wolf-etal-2020-transformers} and PyTorch \citep{pytorch2024} for our implementations.

\subsection{Results}

We present the results in~\Cref{figure:f1_correlation}, where each subplot's legend displays the parameters of the log-linear regression for entity types that overlap between the training dataset and evaluation benchmarks, as well as the average zero-shot F1 score for non-overlapping entity types. We make several observations:

\noindent\textbf{Better performance for overlapping entities.} Evaluation entity types that are also present in the synthetic fine-tuning datasets consistently perform better than those that are absent from the fine-tuning data. However, we note one exception: with LitSet, true zero-shot performance is higher when there are fewer than 100 support examples of an entity type. As \citet{golde-etal-2024-large} explain, this can be attributed to the sparse NER annotations in their dataset, as the original annotations are intended for entity linking rather than named entity recognition.

\noindent\textbf{Better performance for frequent entities.} A second important factor is the number of training instances for overlapping entity types. We observe a positive correlation between the number of entity mentions and the performance of individual entity types across all models. The correlation ranges from $0.04 \log_{10}(x)$ (NERetrieve, AskNews) to $0.08 \log_{10}(x)$ (LitSet), indicating that the benefits of LLM-annotated and automatically derived datasets do not diminish at a fixed point, even though increasingly larger amounts of data are needed for further gains.

\noindent\textbf{Discussion.} Our experiment indicates that overlaps between datasets can indeed inflate zero-shot transfer performance when synthetic data is used. Further, our findings suggest that training datasets generated by LLMs may show significant alignment with existing evaluation benchmarks for NER. 


\section{\metricname{}} \label{sec:metric}
The previous experiments show two key challenges in current zero-shot NER evaluations: (\textit{1}) Overlapping entity types inflate the transfer evaluations of zero-shot models, and (\textit{2}) LLMs may generate ideal datasets for fixed evaluation settings, undermining the concept of low-resource evaluations. Therefore, future evaluations must distinguish between improvements coming from sophisticated datasets and those achieved through new data-efficient approaches that do not depend on overlapping entity types.

To address these challenges, we introduce \metricname{} to quantify label shift between fine-tuning datasets and evaluation benchmarks based on the semantic similarity of the respective entity type sets. \metricname{} considers two key factors: (1) the semantic similarity between evaluation and training entity types, and (2) the support for each training entity type. The core idea is that if the evaluation entity type is ``person'' and the set of training entity types contains a closely related type, such as ``human'', with substantial support, we can expect strong performance. In contrast, if the closest training entity type to "person" is a less related type like ``location'' with limited support, we can expect a worse performance.

To compute semantic similarity, we use a \texttt{sentence-transformer} \citep{reimers-gurevych-2019-sentence} to embed evaluation and training entity types, calculate cosine similarity, and clip negative values to keep the metric within a 0 to 1 range. For the second factor, we introduce a hyperparameter, $K$, which limits the number of support examples considered. In our experiments, we set $K = 1000$, meaning that up to 1000 closest training entity types are considered, measured by their support. We further weight these similarities by a Zipfian distribution \citep{zipf1949}, prioritizing the most similar entity types, as they are likely to have the greatest impact on transfer performance.

\noindent\textbf{Definition.} Let $\mathcal{L}^{\mathcal{D}}$ and $\mathcal{L}^{\mathcal{Z}}$ represent the sets of all entity types in the fine-tuning dataset and the zero-shot benchmarks, respectively. Additionally, let $\mathcal{C}$ denote the set of counts for each entity type $\ell^{\mathcal{D}} \in \mathcal{L}^{\mathcal{D}}$, and let $\theta$ represent the \texttt{all-mpnet-base-v2} sentence-transformer model. For any entity type $\ell^{\mathcal{Z}} \in \mathcal{L}^{\mathcal{Z}}$ from the evaluation benchmarks and any entity type $\ell^{\mathcal{D}} \in \mathcal{L}^{\mathcal{D}}$ from the training dataset, we calculate the clipped cosine similarity as follows:
$$
\varphi_{\text{clip}}(\ell^{\mathcal{Z}}, \ell^{\mathcal{D}}) = \max(\varphi(\theta(\ell^{\mathcal{Z}}), \theta(\ell^{\mathcal{D}})), 0)
$$
\noindent where $\varphi(\cdot, \cdot)$ denotes the standard cosine similarity. We can now calculate the similarity between a given evaluation entity type $\ell^{\mathcal{Z}}$ and all training entity types, resulting in the set:
$$\mathcal{S}^{\ell^{\mathcal{Z}}} = \{\varphi_{\text{clip}}(\ell^{\mathcal{Z}},\ell^{\mathcal{D}}_1), \dots , \varphi_{\text{clip}}(\ell^{\mathcal{Z}},\ell^{\mathcal{D}}_j)\}$$

\noindent We then repeat each element in $\mathcal{S}^{\ell^{\mathcal{Z}}}$ according to the corresponding support $c^{i} \in \mathcal{C}$ for the training entity type $\ell^{\mathcal{D}}_i$ to account for the number of mentions of each training entity type:
$$
repeat(\mathcal{S}^{\ell^{\mathcal{Z}}}, \mathcal{C}) =  \{\underbrace{s^{1},.., s^{1}}_{c^{1}-\text{ times}},.., \underbrace{s^{j},.., s^{j}}_{c^{j}-\text{ times}}\}
$$ 

\noindent with $s^i = \varphi_{\text{clip}}(\ell^{\mathcal{Z}},\ell^{\mathcal{D}}_i)$. We then sort the repeated set of all similarities between the evaluation entity type $\ell^{\mathcal{Z}}$ and all training entity types and select the top-$K$ similarities.
$$
\mathcal{S}^{\ell^{\mathcal{Z}}} = sort(repeat(\mathcal{S}^{\ell^{\mathcal{Z}}}, \mathcal{C}))_{[:K]}
$$

\noindent Once we determined the top-$K$ similarities for evaluation entity type $\ell^{\mathcal{Z}}$, we compute the weighted average using the position $k$ of each similarity value:
$$
\textsc{Familiarity}(\ell^{\mathcal{Z}}) = \frac{\sum_{k = 1}^{K} \mathcal{S}^{\ell^{\mathcal{Z}}}_k \cdot \frac{1}{k}}{\sum_{k = 1}^{K} \frac{1}{k}}
$$

\noindent Finally, we marco-average \metricname{} for each $\ell^{\mathcal{Z}} \in \mathcal{L}^{\mathcal{Z}}$, resulting in an aggregated score for the entire transfer setting.

\noindent To account for the number of mentions of each training entity type $\ell^{\mathcal{D}}_i$, we weight each element in $\mathcal{S}^{\ell^{\mathcal{Z}}}$ by the corresponding probability distribution vector $\mathcal{P}^{\ell^{\mathcal{D}}}$, which represents the relative frequency of each training entity type:

\noindent where $s^i = \varphi_{\text{clip}}(\ell^{\mathcal{Z}}, \ell^{\mathcal{D}}_i)$ and the distribution vector $\mathcal{P}^{\ell^{\mathcal{D}}}$ ensures that entity types with higher mention counts contribute proportionally more to the similarity calculation.

\noindent We then sort the weighted set of all similarities between the evaluation entity type $\ell^{\mathcal{Z}}$ and all training entity types and select the top-$K$ similarities:

\noindent Once we have determined the top-$K$ similarities for evaluation entity type $\ell^{\mathcal{Z}}$, we compute the weighted average using the position $k$ of each similarity value:

\noindent Finally, we macro-average \metricname{} for each $\ell^{\mathcal{Z}} \in \mathcal{L}^{\mathcal{Z}}$, resulting in an aggregated score for the entire transfer setting.

\section{Experiments}
\label{sec:experiments}

We evaluate \metricname{} in various settings to assess its ability to measure label shift in zero-shot NER transfer scenarios. We examine its correlation with traditional transfer performance, the impact of design choices (embedding model and top-$K$ similarities), and how \metricname{} can be used to create NER tasks of varying difficulty.

\subsection{\metricname{} in Current Evaluations}
\noindent\textbf{Setup.}
We reuse the models from~\Cref{sec:validation_experiment} and compute \metricname{} for each setup to evaluate whether our metric correlates with transfer performance of models trained on different synthetic datasets. We report the values of our metric alongside the macro-averaged F1 scores across all seven zero-shot benchmarks, as well as the percentage of overlapping entity types between each training dataset and the combined entity types of all evaluation benchmarks.

\begin{table}
\centering
\normalsize
\begin{tabular}{lccc}
\toprule
Train & \multirow[b]{2}{*}{F1} &  \multirow[b]{2}{*}{\small{Pearson} $r$} & \multirow[b]{2}{*}{\small{\metricname{}}}\\
Dataset $\mathcal{D}$ & & & \\
\midrule
NERetrieve & 0.287 & 0.517 & 0.563 \\
LitSet & 0.380 & 0.340 & 0.695 \\
NuNER & 0.551 & 0.299 & 0.893 \\
PileNER & 0.568 & 0.310 & 0.887 \\
AskNews & 0.585 & 0.457 & 0.899 \\
\bottomrule
\end{tabular}
\caption{Zero-shot F1 scores and \metricname{}, macro-averaged over all seven evaluation benchmarks. \metricname{} quantifies the label shift between fine-tuning and zero-shot benchmarks, explaining why models trained on certain synthetic datasets result in better performance.}
\label{table:results_familarity}
\vspace{-3mm}
\end{table}

\begin{figure*}[ht]
\vspace{-3mm}
\includegraphics[width=\textwidth]{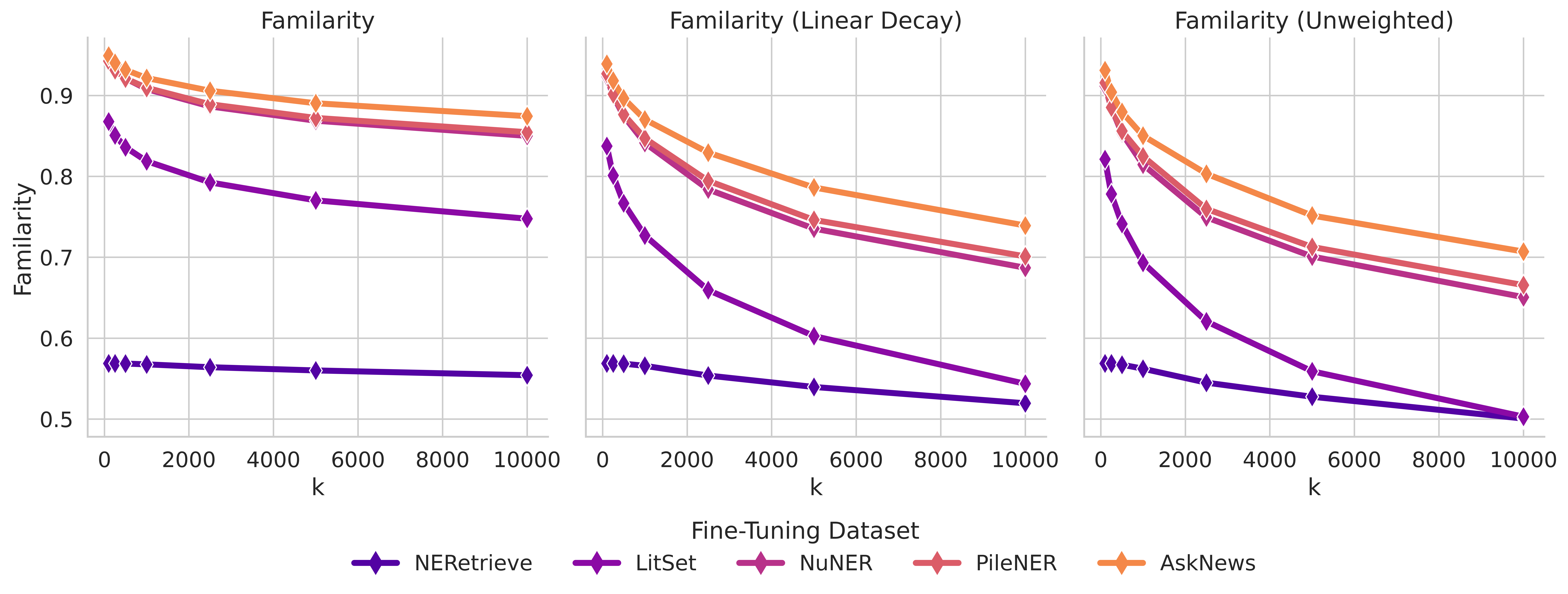}
\vspace{-8mm}
\caption{\metricname{} for different values of $k$ and using different rank weights.}
\vspace{-4mm}
\label{figure:comparison_hyperparameter}
\end{figure*}


\noindent\textbf{Results.} We present the average zero-shot transfer results, Pearson correlation values $r$ (between \metricname{} and F1, macro-averaged over all evaluation entity types), and \metricname{} scores in~\Cref{table:results_familarity}. Our analysis shows that models trained on NuNER, PileNER, and AskNews achieve the highest F1 scores (> 55.0) and high \metricname{} values (> 0.88), suggesting strong alignment between these models and the evaluation entity types. In contrast, the automatically derived datasets, NERetrieve and LitSet, have lower F1 scores (28.7 and 38.0, respectively) and correspondingly lower \metricname{} values (0.563 and 0.695), reflecting a greater label shift between training and evaluation sets. Additionally, the Pearson correlation coefficients (\( r \)) are consistently positive but moderate (0.299–0.517). This suggests that the semantic similarity between entity types in the training and evaluation sets is correlated with transfer performance, though it is not the only factor influencing the final results.

We can summarize that a smaller label shift (similar sets of entity types in training and evaluation datasets) results in higher zero-shot transfer performance. Therefore, considering this factor is crucial for making fair comparisons between different models or architectures in zero-shot NER settings. We further note that \metricname{} complements existing metrics like F1 by making the impact of entity type overlaps explicitly visible, leading to a more interpretable comparison.

\subsection{Impact of $K$} \label{sec:hyperparameter}eIn this experiment, we explore the effect of the hyperparameter $K$, which controls how many entity types (measured by their support) are considered when computing \metricname{} for a given evaluation entity type. Thus, $K$ can be seen as the number of support examples from which we expect a model to learn a specific entity concept. We recall that we use $K = 1000$ to include not only the closest types but also a variety of similar types that may help in learning the class definition of certain entity types.

\noindent\textbf{Setup.}
We reuse the models trained in~\Cref{sec:validation_experiment} and recompute \metricname{} using various values of $K$, ranging from 100 to 10,000. Additionally, we compare our default Zipfian weighting with two other approaches: linear decay ($p(k) = \frac{|K| - k}{|K|}$), which gradually reduces the influence of lower-ranked entity types, and an unweighted approach, which treats all entity types equally. This comparison helps us understand how different weighting strategies interact with $K$ and influence \metricname{} scores.

\noindent\textbf{Results.}
We present the results in~\Cref{figure:comparison_hyperparameter}. We observe that \metricname{} values are higher for smaller values of $K$ and decrease as $K$ increases. This is expected, as smaller $K$ values emphasize entity types most similar to the evaluation types, while larger $K$ values incorporate more distant, less similar types. In particular, the unweighted results reveal that most datasets have a few highly similar entity types, but the similarity declines rapidly beyond those. Applying weighting schemes such as linear decay or Zipf smooths this decline, which is desirable because it makes \metricname{} less sensitive to variations in $K$. Crucially, the relative ranking of datasets remains stable across different values of $K$ and weighting methods. Based on these observations, we argue that the optimal configuration for \metricname{} uses $K = 1000$ with Zipf weighting.

\subsection{Different Embedding Models} \label{sec:different_transformers}

\begin{table*}[ht]
\vspace{-3mm}
\centering
\begin{tabular}{lccccc}
\toprule
 & \multicolumn{5}{c}{Fine-Tuning on:} \\
 & NERetrieve & LitSet & NuNER & PileNER & AskNews \\
\midrule
\diameter~Zero-Shot F1 & 28.7 & 38.0 & 55.1 & 56.8 & 58.5 \\
\midrule
fasttext-crawl-300d-2M & \underline{0.595} & \underline{0.712} & 0.898 & 0.857 & 0.874 \\
fasttext-wiki-news-300d-1M & \underline{0.718} & \underline{0.791} & 0.920 & 0.892 & 0.904 \\
glove-6B-300d & \underline{0.654} & \underline{0.743} & 0.910 & 0.879 & 0.893 \\
\midrule
bert-base-uncased & \underline{0.823} & \underline{0.872} & 0.956 & 0.948 & 0.954 \\
distilbert-base-uncased & \underline{0.883} & \underline{0.917} & 0.973 & 0.968 & 0.972 \\
\midrule
all-mpnet-base-v2 & \underline{0.563} & \underline{0.695} & 0.893 & 0.887 & \underline{0.899} \\
all-miniLM-L6-v2 & \underline{0.605} & \underline{0.701} & 0.901 & 0.893 & \underline{0.905} \\
\bottomrule
\end{tabular}
\caption{\metricname{} using different embedding models. Underscored values indicate cases where \metricname{} matches the ranking of the macro-averaged F1 score.}
\label{table:different_transformers_results}
\vspace{-3mm}
\end{table*}

Another important hyperparameter is the embedding model $\theta$. In this experiment, we examine how the choice of embedding model affects the values of \metricname{} and the potential impact on our metric's outcomes.

\noindent\textbf{Setup.}
We reuse the models trained in~\Cref{sec:validation_experiment} but change the underlying embedding model to compute \metricname{}. One potential limitation of transformers is that they encode tokens in context, which may be less effective for short entity type descriptions, often consisting of single words. Therefore, we compare our chosen model with standard transformers, additional sentence-transformers, and classical word embeddings. Specifically, we consider:

\noindent\textbf{Classical Word Embeddings:} We include two fasttext models \citep{bojanowski-etal-2017-enriching}, \texttt{fasttext-crawl-300d-2M} and \texttt{fasttext-wiki-news-300d-1M}, along with the largest GloVe embedding \citep{pennington-etal-2014-glove}, \texttt{glove-6B-300d}.

\noindent\textbf{Classical Transformers:} We include two widely used transformers: \texttt{bert-base-uncased} \citep{devlin-etal-2019-bert} and \texttt{distilbert-base-uncased} \citep{sanh2020distilbertdistilledversionbert}, which are not specifically trained for semantic similarity measurement.

\noindent\textbf{Sentence Transformers:} We compare the selected \texttt{all-mpnet-base-v2} with another sentence-transformer model, \texttt{all-miniLM-L6-v2} \citep{reimers-gurevych-2019-sentence}.

\noindent\textbf{Results.}
We present results in~\Cref{table:different_transformers_results}. First, all embedding models show similar trends: low-performing models, such as those trained on NERetrieve or LitSet, consistently achieve the lowest similarity scores across all embedding models. For high-performing models (NuNER, PileNER, and AskNews), all embedding models provide reasonable results, with high F1 scores and \metricname{} values, accurately reflecting the overall low label shift. Despite the small absolute differences, \metricname{} remains close across our trained models, capturing the overall label shift effectively.

Our results indicate that \metricname{} performs well with various embedding models. However, the choice of embedding model affects the scale of similarity scores: classical transformer models tend to consistently produce high \metricname{} scores (> 82.3) across all settings, which is not ideal. We are interested in an embedding model that can clearly distinguish between different label shifts. We argue that classical word embeddings, particularly \texttt{fasttext-crawl-300d-2M}, and the \texttt{all-mpnet-base-v2} sentence-transformer perform best in this regard. Given that label descriptions may become more detailed with future synthetic datasets, we argue using \texttt{all-mpnet-base-v2} is the best option. However, if computational efficiency is a priority, \texttt{fasttext-crawl-300d-2M} is a viable alternative.

\subsection{Using \metricname{} to Generate Training Splits of Varying Difficulty} \label{sec:analysis_trust}

In this section, we explore how \metricname{} can be applied to create training splits (subsets of the original datasets) with varying levels of difficulty. If \metricname{} effectively captures and explains label shift in NER transfer settings, it should enable us to generate splits with either low or high label shifts accordingly.

\noindent\textbf{Setup.} We create a similarity matrix $\mathcal{M}$ using our embedding model $\theta$ containing the similarities between each pair of training entity type $\ell^{\mathcal{D}} \in \mathcal{L}^{D}$ and evaluation entity type $\ell^{\mathcal{Z}} \in \mathcal{L}^{Z}$:
$$
\mathcal{M}_{ij} = \varphi_{\text{clip}}(\theta(\ell^{\mathcal{D}}), \theta(\ell^{\mathcal{Z}}))
$$

\noindent such that $\mathcal{M} \in \mathbb{R}^{|\mathcal{L}^{D}| \times |\mathcal{L}^{Z}|}$. We assign a single value to each training label (row of $\mathcal{M}$) by either (1) taking the maximum similarity or (2) computing the entropy over all evaluation labels, which indicates how well a training entity type aligns with the evaluation entity type set. Based on this, we create training splits with low, random, or high label shifts by selecting training entity types according to quantiles of $\mathcal{M}$. For example, the top 1\% quantile in the maximum similarity matrix $\mathcal{M}$ includes training entity types that are highly similar to at least one evaluation entity type. A split consisting solely of these entity types would result in a training split with low label shift. Details of the selection process are provided in~\Cref{sec:distribution_and_selection_ablation}.

For these experiments, we use NuNER and PileNER, as they show the best performance and are standalone datasets (unlike AskNews, which requires a pre-fine-tuned model). For each dataset, we filter it to include only entity types with low, medium, or high label shifts, removing all others. We then train models as described in previous sections, but for 10,000 steps instead of 60,000, as the filtered subsets are significantly smaller than the original datasets, reducing the risk of overfitting.

\begin{table}
\centering
\begin{tabular}{lllcc}
\toprule
$\mathcal{D}$ & Agg. & \small{Label Shift} & \small{\metricname{}} & F1 \\
\midrule
\multirow[c]{6}{*}{{\rotatebox{90}{NuNER}}} & \multirow[c]{3}{*}{Entropy} & low & 0.806 & 45.8 \\
 &  & medium & 0.630 & 33.6 \\
 &  & high & 0.530 & 28.0 \\
\cline{2-5}
 & \multirow[c]{3}{*}{Max. $\varphi$} & low & 0.865 & 42.5 \\
 &  & medium & 0.637 & 30.9 \\
 &  & high & 0.364 & 23.7 \\
\cline{1-5} \cline{2-5}
\multirow[c]{6}{*}{{\rotatebox{90}{PileNER}}} & \multirow[c]{3}{*}{Entropy} & low & 0.880 & 45.8 \\
 &  & medium & 0.534 & 30.8 \\
 &  & high & 0.596 & 33.3 \\
\cline{2-5}
 & \multirow[c]{3}{*}{Max. $\varphi$} & low & 0.896 & 43.5 \\
 &  & medium & 0.551 & 26.2 \\
 &  & high & 0.389 & 29.8 \\
\bottomrule
\end{tabular}
\caption{Using \metricname{}, we generate subsets of PileNER and NuNER with varying levels of difficulty. These splits can be produced using either entropy-based selection or maximum similarity-based selection.}
\label{table:results_new_splits}
\vspace{-3mm}
\end{table}

\noindent\textbf{Results.} 
The results in~\Cref{table:results_new_splits} show that \metricname{} can successfully create training splits of varying difficulty, regardless of the aggregation method (entropy or maximum similarity). Models trained on splits with low label shifts consistently achieve higher \metricname{} values and F1 scores, indicating better alignment with the evaluation data. For instance, in the low label shift setting for NuNER with entropy aggregation, \metricname{} reaches 0.806 and the F1 score is 45.8, whereas in the high label shift setting, these values drop to 0.530 and 28.0, respectively. Similarly, for PileNER, the F1 score decreases by 17.8 points between the low and high label shift settings using entropy aggregation.

Interestingly, entropy aggregation yields better results in low label shift settings compared to maximum similarity, while maximum similarity produces lower scores in high label shift settings. This suggests that entropy aggregation is more effective for capturing low label shift, whereas maximum similarity is better suited for generating high label shift splits.

\section{Related Work}

The problem of NER can be formulated in many ways such as span classification \citep{yu-etal-2020-named}, question answering \citep{li-etal-2020-unified}, and text generation \citep{cui-etal-2021-template,ma-etal-2022-template}. The emergence of large language models has recently transformed many downstream NLP tasks through natural language prompting \citep{min-etal-2022-rethinking,dong-2023-survey}, including NER \citep{aly-etal-2021-leveraging,dozen2021nguyen,li-etal-2022-prompt-based-text,ma-etal-2022-label,chen-etal-2023-prompt,shen-etal-2023-promptner}. Our work contributes to this line of research by measuring the label shift of entity type prompts.

\noindent
\textbf{Similarity Metrics.} Many works exist on evaluating outputs generated by a model with the target sequence using similarity metrics such as BERTscore \citep{zhang-2020-bertscore}, BARTscore \citep{yuan-2021-bartscore}, or SEMscore \citep{aynetdinov-2024-semscore} as well as task-specific similarity metrics such as SEM-F1 \citep{bansal-etal-2022-sem} or SAS \citep{risch-etal-2021-semantic}. We follow this idea by comparing the semantic similarity between fine-tuning and zero-shot entity types.

\noindent
\textbf{Zero-Shot NER.} We have recently observed increasingly capable NER systems trained on large-scale datasets \citep{wang-2023-instructuie,lou-2023-universalie,zhou-2023-universalner,sainz-2024-gollie}. These works stand out because they have been fine-tuned on datasets covering thousands of entity types. Considering the progress of LLMs, we expect more contributions generating tailored datasets \citep{schick-schutze-2021-generating,ye-etal-2022-zerogen,ye-etal-2022-progen,li-etal-2023-synthetic} for downstream tasks. Our work supports this line of research to better evaluate future contributions by explicitly measuring the label shift.

\section{Conclusion}
This paper explores how the label shift between synthetically produced training datasets affects the performance of zero-shot NER as evaluated in current benchmark scnearios. As LLMs advance, creating improved datasets that align with the chosen zero-shot benchmarks to enhance transfer performance becomes more accessible. As a consequence, evaluation settings become less comparable. Thus, we introduce \metricname{} to quantify the connection between fine-tuning and zero-shot datasets and show how it can achieve fairer comparisons. Although the automatic generation of datasets holds promise for future NER research, it is crucial to foster data-efficient research by conducting zero-shot NER in scenarios where fine-tuning datasets do not contain closely related entity types.

To enable the research community to efficiently compute \metricname{} and incorporate it into future research, we make all code publicly available as open source. Further, we publish three benchmark scenarios for different levels of transfer difficulty to aid researchers in fine-grained analysis of zero-shot NER.


\section*{Limitations}

\metricname{} is specifically designed for transfer settings in the NER domain, but addresses a broader issue: label shift in transfer learning. Although we validated our metric only for NER, it is possible - if not likely - that the metric could yield different results when applied to other downstream tasks.

Furthermore, our \metricname{} metric is designed for models trained from scratch and does not account for the extensive pre-training of LLMs. Since pre-trained models may already contain implicit knowledge of certain entities and phrases, such as ``Google is a technology company,'' our method does not currently measure the impact of such prior knowledge. Future work could explore complementary evaluation techniques to assess the impact of pre-training more accurately.

Our metric is designed for datasets that contain precise and clearly defined entity types, which is especially important in the context of the increasing use of synthetic datasets. Synthetic datasets often leverage structured knowledge bases and large language models to generate fine-grained entity labels. However, the reliance of the metric on such detailed annotations means that it is less effective when applied to simpler, high-resource datasets where multiple concepts might be grouped into a single broad entity class. For example, in datasets where a general category like ``organization'' encompasses various subtypes (e.g., companies, non-profits and government agencies), \metricname{} may not accurately capture the true difficulty of transfer learning. This limitation suggests that the metric is best suited for evaluations where entity types are well-defined and separated, rather than for datasets where broad classes mask underlying distinctions.

Additionally, our metric does not account for the actual context in which entity mentions occur, which can significantly impact final model performance, especially in the presence of label noise. \metricname{} measures semantic similarity between entity types based on their descriptions or definitions, but it does not evaluate how these entities are annotated in practice within the training and evaluation datasets. As a result, the metric might yield a high similarity score when entity types appear closely related based on their definitions, even if the actual annotations differ considerably in context. For instance, two entity types might be semantically similar (e.g., ``artist'' and ``musician''), but if one dataset consistently annotates "musician" while another uses "artist" for the same context, the differing annotation standards could lead to performance inconsistencies. This discrepancy means that while \metricname{} offers insight into type overlap, it may not fully capture the practical challenges of adapting to label noise and annotation inconsistencies during model evaluation.

\section*{Acknowledgments}
We thank all reviewers for their valuable comments. Jonas Golde is supported by the Bundesministerium für Bildung und Forschung (BMBF) as part of the project ``FewTuRe'' (project number 01IS24020). Alan Akbik and Patrick Haller are supported by the Deutsche Forschungsgemeinschaft (DFG, German Research Foundation) under Emmy Noether grant ``Eidetic Representations of Natural Language'' (project number 448414230). Further, Alan Akbik and Max Ploner are supported under Germany’s Excellence Strategy ``Science of Intelligence'' (EXC 2002/1, project number 390523135). Fabio Barth is supported by the Bundesministerium für Wirtschaft und Energie (BMWi) as part of the project ``OpenGPT-X'' (project number 68GX21007D).

\bibliography{anthology_1,custom}

\clearpage
\appendix

\section*{Appendix}

\section{Detailed Results} \label{sec:detailed_results}

The results in~\Cref{table:detailed_zero_shot_results} compare the zero-shot transfer performance of all trained models and benchmarks considered. Overall, AskNews achieves the highest average performance (58.5), demonstrating strong results in most benchmarks, including top scores in AI (57.0) and Science (65.9). PileNER closely follows with an average score of 56.8, excelling particularly in Politics (70.7), Literature (61.3), and Music (68.1). NuNER also performs well, achieving an average score of 55.1, with consistent performance across most domains, including a strong result in Science (57.4). In contrast, LitSet and NERetrieve achieve lower average scores, with 38.0 and 28.7, respectively. NERetrieve shows weaker performance across all benchmarks, especially in the Restaurant domain (16.8). These results highlight the variability in transfer performance depending on the fine-tuning dataset, with datasets like AskNews and PileNER generally providing more robust coverage across diverse domains compared to LitSet and NERetrieve.

\begin{figure*}[ht]
\includegraphics[width=\textwidth]{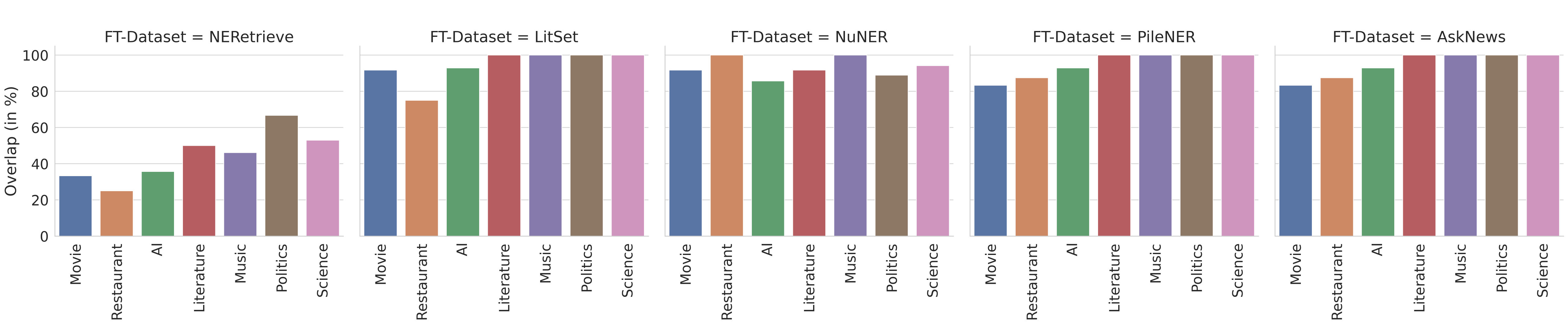}
\caption{Overlapping entity types between considered synthetic training datasets and all evaluation benchmarks.}
\label{figure:overlaps_overview}
\end{figure*}

Further,~\Cref{figure:overlaps_overview} illustrates the overlap between entity types present in all considered fine-tuning datasets and those in the evaluation benchmarks. We simply measures whether each entity type in the benchmarks is also found in the fine-tuning datasets. NERetrieve displays notably low scores, indicating that it lacks many of the entity types present in the evaluation benchmarks. In contrast, the other datasets—NuNER, PileNER, LitSet, and AskNews—show high overlap scores, with values exceeding 80\% and reaching up to 100\%. This suggests that these datasets contain all or nearly all the entity types considered in the benchmarks. However, despite this high overlap, our experiments highlight the importance of considering the semantic similarity and the amount of entity mentions for each entity type. For example, LitSet, despite having a high overlap, performs worse than NuNER, PileNER, and AskNews. This result emphasizes that merely having the same entity types is insufficient; the quality and contextual understanding of those types matter. Additionally, the figure reinforces that no benchmark can be considered truly zero-shot, as all show significant overlap with the fine-tuning datasets.

\begin{table*}[ht]
\centering
\begin{tabular}{l|ccccccc|c}
\toprule
FT-Dataset & Movie & Restaurant & AI & Science & Politics & Literature & Music & Average \\
\midrule
NERetrieve & 35.8 & 16.8 & 24.2 & 34.6 & 28.5 & 27.1 & 34.1 & 28.7 \\
LitSet & 46.9 & 29.3 & 33.8 & 31.2 & 43.6 & 47.2 & 34.2 & 38.0 \\
NuNER & 43.7 & \textbf{46.7} & 47.7 & 57.4 & 64.5 & 59.9 & 65.5 & 55.1 \\
PileNER & 51.0 & 36.4 & 52.6 & 57.3 & \textbf{70.7} & \textbf{61.3} & \textbf{68.1} & 56.8 \\
AskNews & \textbf{56.6} & 41.8 & \textbf{57.0} & \textbf{65.9} & 62.9 & 60.4 & 65.2 & \textbf{58.5} \\
\bottomrule
\end{tabular}
\caption{Transfer results for each evaluation benchmark considered. Results are averaged over three different seeds.}
\label{table:detailed_zero_shot_results}
\end{table*}

\section{Creating Splits of Varying Difficulty using \metricname{}} \label{sec:distribution_and_selection_ablation}
We compute a similarity matrix $\mathcal{M}$ where each row represents a training entity type from $\mathcal{L}^{D}$ and each column represents an evaluation entity type from $\mathcal{L}^{\mathcal{Z}}$. To aggregate the similarity scores for each training entity type $\ell^{\mathcal{D}}$, we apply two different strategies: 

\begin{description}[leftmargin=0cm]
    \item [Maximum Similarity Selection.] For each row $i$, we take the maximum similarity score across all columns $j$, which captures the highest similarity between a training entity type $\ell^{\mathcal{D}}_i$ and any evaluation entity type $\ell^{\mathcal{Z}}_j$:
    \[
    \mathcal{M}^{\max}_{i} = \max_{j} \mathcal{M}_{ij}, \quad \forall \, i \in \{1, \ldots, |\mathcal{L}^{D}|\}.
    \]

    \item[Entropy-Based Selection.] For each row $i$, we calculate the entropy over the similarity values to measure how evenly distributed the similarities are across all evaluation entity types. Lower entropy indicates that the similarities are concentrated around one or a few evaluation types, while higher entropy suggests a more uniform distribution:
    \[
    \mathcal{M}^{\text{ent}}_{i} = - \sum_{j=1}^{|\mathcal{L}^{Z}|} p_{ij} \log(p_{ij}), \quad \forall \, i \in \{1, \ldots, |\mathcal{L}^{D}|\},
    \]
    where the probability $p_{ij}$ is defined as:
    \[
    p_{ij} = \frac{\exp\left(\frac{\mathcal{M}_{ij}}{T}\right)}{\sum_{j=1}^{|\mathcal{L}^{Z}|} \exp\left(\frac{\mathcal{M}_{ij}}{T}\right)}, \quad T = 0.01.
    \]
    The low temperature value ($T = 0.01$) forces the distribution to peak around the highest similarity scores, emphasizing the most meaningful alignments between training and evaluation types.
\end{description}

After aggregating, the resulting scores $\mathcal{M}^{\max}$ and $\mathcal{M}^{\text{ent}}$ are in $\mathbb{R}^{|\mathcal{L}^{D}|}$, representing the relevance for each training entity type considering the entire evaluation entity types.

In the subsequent analysis, we select quantiles from the aggregated scores:
\begin{itemize}
    \item For $\mathcal{M}^{\max}$, we select the top 1\% of similarity values to represent the low label shift transfer setting, as these training entity types exhibit the highest similarity to any evaluation entity type. Conversely, the lowest 1\% of scores correspond to a high label shift transfer setting, as these training types have low similarity to all evaluation entity types.
    \item Conversely, for $\mathcal{M}^{\text{ent}}$, we select the lowest 1\% of entropy scores for the low label shift transfer setting, indicating training entity types that have a concentrated similarity with one or a few evaluation labels. The top 1\% represent the high label shift transfer setting, as these scores reflect a uniform distribution over all evaluation entity types.
\end{itemize}

By using these quantile selections, we can distinguish between training entity types that are more likely to yield better performance given the evaluation types and those that are presumably less suitable for the evaluation entity types.

\noindent\textbf{Quantile Selection.} The quantile selection for generating training splits is adapted based on both the training dataset and the metric used, taking into account the number of labels in each dataset.For the maximum similarity-based selection:
\begin{itemize}
    \item We focus on the highest quantiles for the low label shift setting and on the lowest quantiles for the high label shift setting, as higher similarity scores indicate closer alignment between training and evaluation entity types.
    \item For PileNER, we select the low 5\% quantile for the high label shift setting and the top 99\% quantile for the low label shift setting.
    \item For NuNER, we use the low 0.5\% quantile for the high label shift setting and the top 99.5\% quantile for the low label shift setting.
\end{itemize}

For the entropy-based selection:
\begin{itemize}
    \item We focus on the lowest quantiles for the low label shift setting and the highest quantiles for the high label shift setting. This is because a lower entropy score indicates that the similarity between the training and evaluation entity types is concentrated around a few specific evaluation types, indicating the training label is valuable for training.
    \item For PileNER, which contains around 15,000 labels, we select the low 1\% quantile for the low label shift setting and top 95\% quantile for the high label shift setting. This broader range is chosen due to the relatively smaller number of labels.
    \item For NuNER, which has over 190,000 labels, we select the low 0.5\% quantile for the low label shift setting and top 99.5\% quantile for the high label shift setting. This narrower selection focuses only on the most highly relevant or irrelevant labels, ensuring that we do not include too many labels in the training split.
\end{itemize}

\begin{figure*}[ht]
\centering
\includegraphics[width=0.7\textwidth]{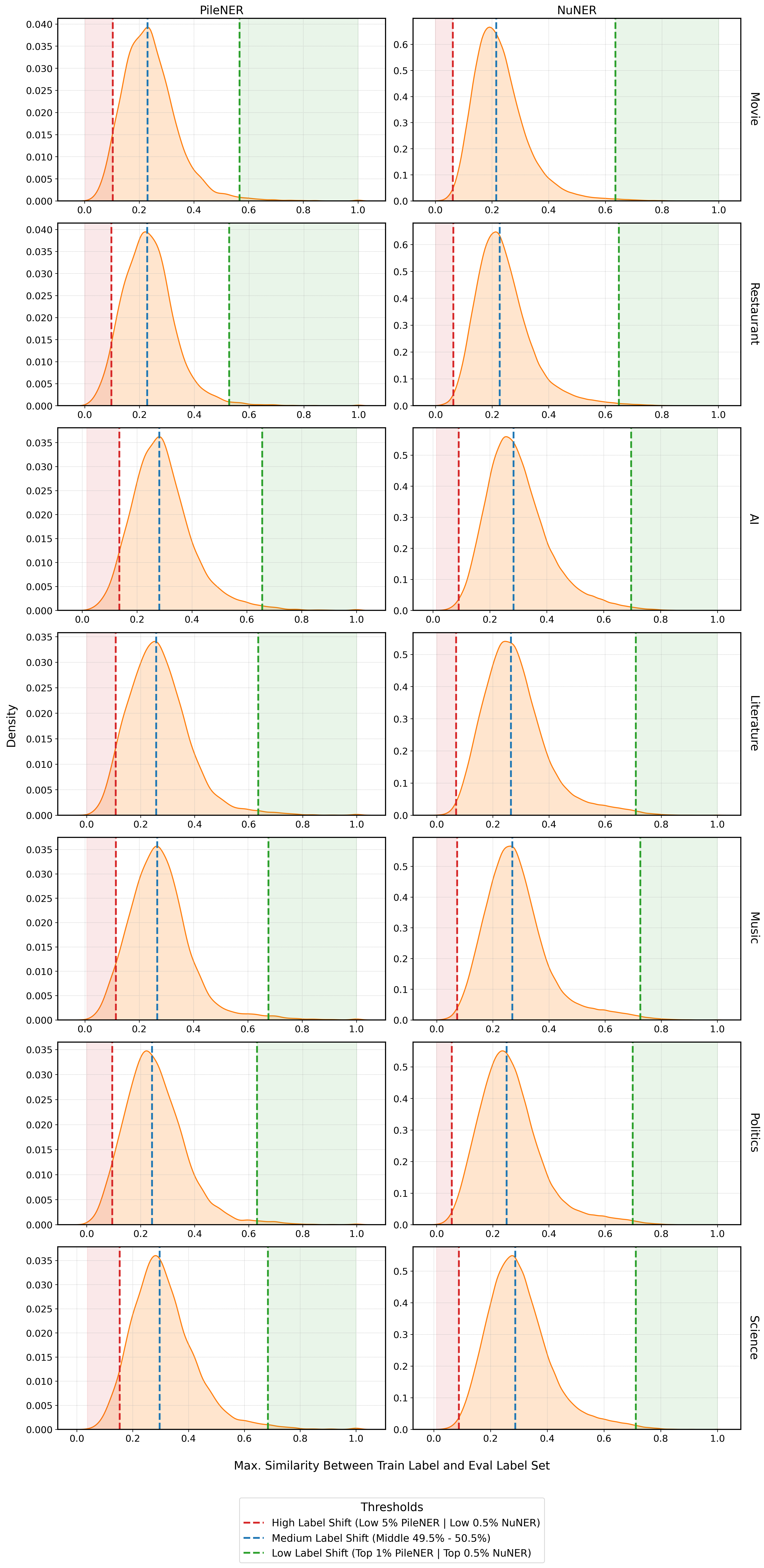}
\caption{Distribution of maximum similarities between all fine-tuning datasets and evaluation benchmarks. Entity types selected for the high label shift setting are indicated in red, those for the label shift setting in blue, and those for the low label shift setting in green.}
\label{figure:distribution_and_selection_ablation}
\end{figure*}

Further, we consider the medium label shift setting to be the 49.5\% - 50.5\% quantile, independent of the dataset. We show an overview of the distribution of max. similarity scores in~\Cref{figure:distribution_and_selection_ablation} and indicate the quantile selection.

\end{document}